\documentclass[letterpaper, 10 pt, conference]{ieeeconf}  

\IEEEoverridecommandlockouts                              
\usepackage{graphics}   
\usepackage{epsfig}     
\usepackage{mathptmx}   
\usepackage{times}      
\usepackage{amsmath}    
\usepackage{amssymb}    
\usepackage{booktabs}   
\usepackage{multirow}   
\usepackage{footnote}
\usepackage{color}
\usepackage{makecell}
\usepackage{cite} 
\makeatletter
\let\NAT@parse\undefined
\makeatother
\usepackage{hyperref}

\title{
    \LARGE \bf
    A Plug-and-Play Learning-based IMU Bias Factor for Robust Visual-Inertial Odometry
}

\author{
    Yang Yi, Kunqing Wang, Jinpu Zhang, Zhen Tan, Xiangke Wang, Hui Shen$^{*}$ and Dewen Hu
    \thanks{
        Y. Yi, K. Wang, J. Zhang, Z. Tan, X. Wang, H. Shen and D. Hu are with the College of Intelligence Science and Technology, National University of Defense Technology, China.
    }
    \thanks{
        * indicates corresponding authors: H. Shen(shenhui@nudt.edu.cn)
    }
    \thanks{
        This work was supported in part by the Science and Technology Innovation Program of Hunan Province (2024QK2006).
    }
}

\begin{document}
\maketitle
\thispagestyle{empty}
\pagestyle{empty}

\begin{abstract}
    Accurate and reliable estimation of biases of low-cost Inertial Measurement Units (IMU) is a key factor to maintain the resilience of Visual-Inertial Odometry (VIO), particularly when visual tracking fails in challenging areas. In such cases, bias estimates from the VIO can deviate significantly from the real values because of the insufficient or erroneous vision features, compromising both localization accuracy and system stability.
    To address this challenge, we propose a novel plug-and-play module featuring the Inertial Prior Network (IPNet), which infers an IMU bias prior by implicitly capturing the motion characteristics of specific platforms. The core idea is inspired intuitively by the observation that different platforms exhibit distinctive motion patterns, while the integration of low-cost IMU measurements suffers from unbounded error that quickly accumulates over time. Therefore, these specific motion patterns can be exploited to infer the underlying IMU bias.
    In this work, we first directly infer the biases prior only using the raw IMU data using a sliding window approach, eliminating the dependency on recursive bias estimation combining visual features, thus effectively preventing error propagation in challenging areas. 
    Moreover, to compensate for the lack of ground-truth bias in most visual-inertial datasets, we further introduce an iterative method to compute the mean per-sequence IMU bias for network training and release it to benefit society. 
    The framework is trained and evaluated separately on two public datasets and a self-collected dataset. Extensive experiments show that our method significantly improves localization precision and robustness. On the EuRoC and TumVi datasets, ATE-RMSE improves by 46\% on average, and RPE-RMSE improves by 48\% on average. On the In-house dataset(indoor), ATE-RMSE also improves by 46\% on average. Furthermore, we apply the model trained in the indoor environment to outdoor scenes and perform testing. The experimental results demonstrate that the model can be successfully applied to outdoor autonomous vehicle. To benefit society, we will open-source our code and release the IMU dataset upon acceptance.
    \end{abstract}
    
    
    
    \section{Introduction}
    High-precision and robust localization technologies are essential for robotics~\cite{panigrahi2022localization,li2024hcto} and augmented reality applications~\cite{dargan2023augmented}. With the miniaturization and cost reduction of visual sensors and Inertial Measurement Units (IMU), Visual-Inertial Odometry (VIO) has emerged as a leading solution for localization. However, in challenging scenarios where visual data is compromised (such as motion blur, illumination changing, or low-texture environments), the reliability of IMU bias estimation for the low-cost devices becomes critical to VIO performance~\cite{li2013high}. More specifically, in the absence of sufficient visual features, IMU bias estimates from the existing VIO systems can deviate significantly from the true values due to the backend optimization of VIO, leading to a degradation in both localization accuracy and system resilience.
    
    \begin{figure}[!htbp]
        \centering
        \includegraphics[width=0.45\textwidth]{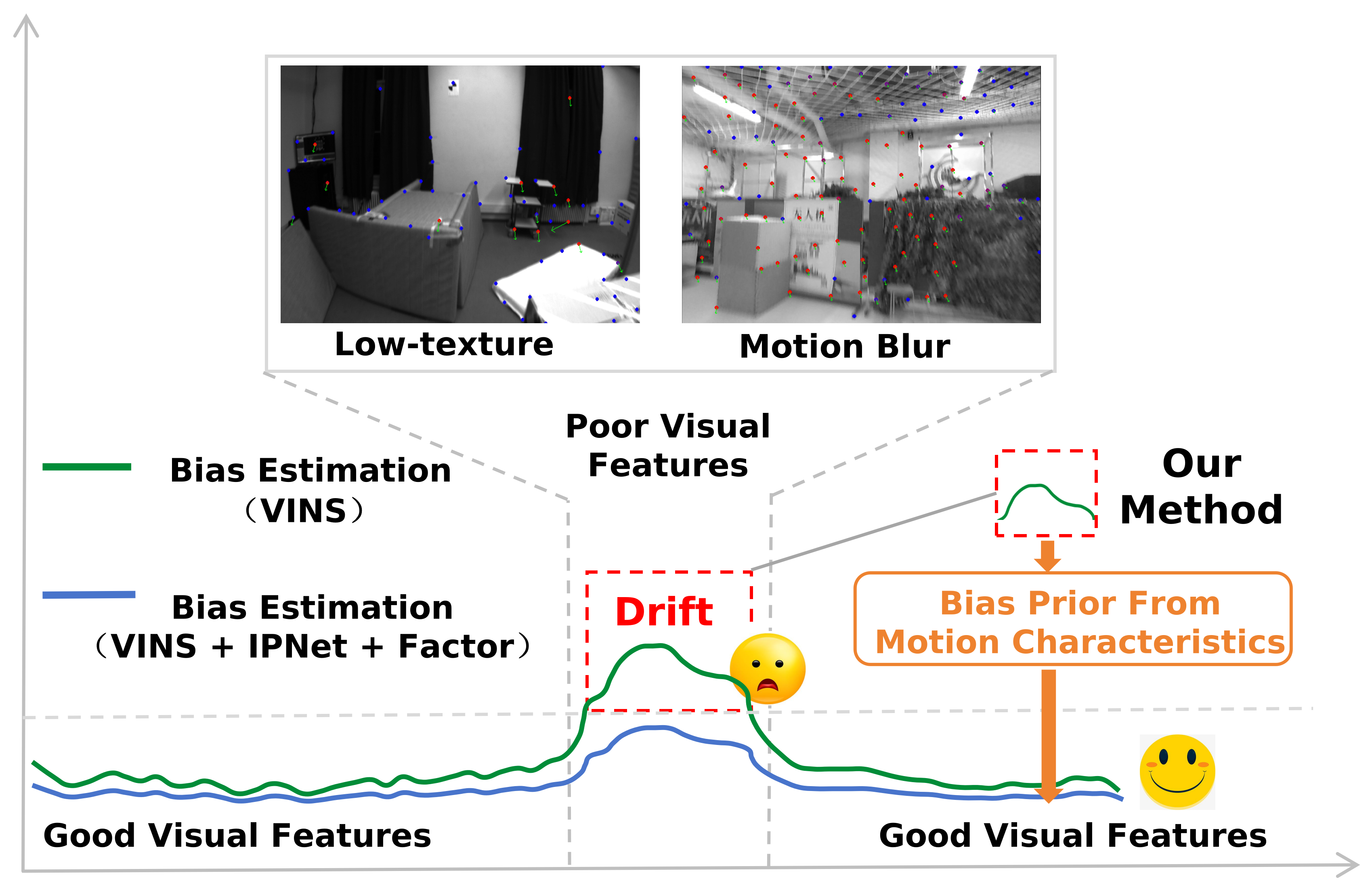}
        \caption{Our method ensures that the estimated IMU bias is more physically consistent, particularly when visual errors are significant, thereby mitigating the degradation of localization precision and system robustness.}
        \label{cover}
    \end{figure}
    
    Existing methods for IMU bias estimation in optimization-based VIO in challenging environments face significant limitations. For instance, the baseline approach using monocular and inertial data in VINS-Fusion \cite{qin2019general} struggles with unbounded adjustments of the bias during optimization, often forcing the bias to fit inaccurate visual constraints caused by motion blur or unbalanced features. HCTO \cite{li2024hcto} attempts to improve the accuracy of low-cost IMU considering human motion patterns, but this method can not be directly applied to other platforms. To address these issues, researchers have turned to deep learning methods to model the dynamic evolution of IMU biases. Similar to our method, HelmetPoser \cite{li2024helmetposer} uses a data-driven approach to estimate head motion using IMU data only, but the pure IMU odometry still faces large drift in diverse scenes. Unlike the recursive framework in \cite{Deep_IMU_2023}, which estimates time-varying bias sequentially and is therefore sensitive to initialization errors and prone to long-term drift, our non-recursive design infers a stable bias prior directly from raw data.  This fundamentally prevents error propagation and improves robustness against label noise by relying on per-sequence mean bias supervision instead of time-varying bias estimation.
    
    \begin{figure*}[htbp]
        \centering
        \includegraphics[width=1.0\textwidth]{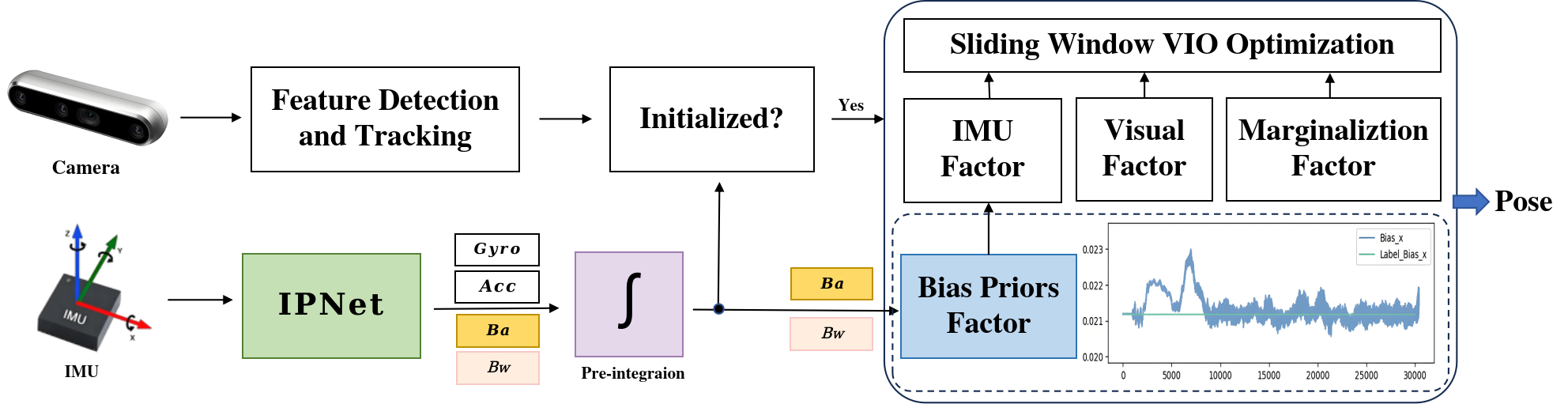}
        \caption{The overview of the proposed system. Given image and IMU data, the framework outputs real-time pose information. At first, the IMU data is passed through the IPNet to estimate the current bias prior. Subsequently, the data stream is divided into two parts: one part inputs the bias prior along with the raw measurements into the pre-integration module, and performs optimization in the backend using IMU factors; the other part incorporates it as a prior factor into the factor graph, imposing reasonable prior constraints on the bias optimization process to ensure the robustness of the bias estimation.}
        \label{frame}
    \end{figure*}
    
    From the perspective of traditional inertial navigation theory, the observability of IMU bias strictly relies on external observation information (such as vision, GPS, etc.), and pure IMU data alone cannot provide sufficient observation conditions\cite{furgale2013unified,martinelli2011vision,hesch2014camera}. However, it is worth noting that certain platform follows specific motion patterns\cite{Deep_IMU_2023}, and data-driven neural networks can capture the implicit relationship between IMU bias and motion characteristics, thereby overcoming the limitations of traditional observability frameworks. Therefore, we propose the Inertial Prior Network (IPNet), which learns a stable bias prior—the most probable bias configuration consistent with the platform’s motion dynamics. IPNet infers this prior directly from raw IMU data using a non-recursive architecture, thereby eliminating the accumulation and propagation of historical errors in visually challenging environments. The output serves as a prior factor injected into the VIO backend, where it collaborates with visual observations through a dynamic constraint mechanism for joint optimization, ensuring the physical plausibility of the bias estimates and significantly improving robustness and positioning accuracy. Furthermore, since the backend optimization can compensate for bias estimate residuals, we simplify the training process by using the average bias of each sequence (in visual feature-rich environments) as the supervision signal, and independently train and evaluate on different datasets. 
    Finally, addressing the common issue in visual-inertial datasets of lacking true bias values, we propose an iterative optimization method based on pose ground truth to compute the sequence's average IMU biases, which is then directly used as the training label for the network. Overall, our main contributions are as follows:
    
    (i) We propose IPNet, a novel network for real-time IMU bias prior inference from raw data, eliminating recursive prediction dependency and preventing error propagation in the visual challenging environment. By constructing bias prior as a plug-and-play prior factor integrated into the VIO framework, the approach significantly enhances localization precision and resilience.
    
    
    (ii) We introduce an iterative method to calculate the average IMU bias for each sequence, addressing the issue of the lack of ground truth in public datasets. The generated average IMU biases are treated as training labels and successfully validated as reliable resources for improving the localization accuracy. The average IMU bias generation process and related data will be released to benefit the community.
    
    (iii) Extensive experiments on public and self-collected multi-platform datasets (handheld and autonomous vehicle) demonstrate significant improvements in localization precision and robustness.
    
    
    The remainder of this paper is organized as follows.
    Section 2 reviews related studies on VIO and learning-based inertial estimation.
    Section 3 introduces the proposed IPNet framework in detail, including its network design, bias prior factor formulation, and the iterative label generation method.
    Section 4 presents the experimental setup, datasets, evaluation metrics, and quantitative results compared to baseline and state-of-the-art methods.
    Section 5 discusses the results, including inference speed, generalization ability, and the requirement of Vicon.  
    Finally, Section 6 concludes the paper and discusses future research directions.
    
    \section{Related work}
    
    \subsection{Visual inertial odometry}
    
    The VIO algorithm was initially developed using the Extended Kalman Filter (EKF), primarily due to the computational limitations of early systems. This line of research has continued to evolve, with notable contributions including ROVIO \cite{ROVIO2015}, MSCKF \cite{MSCKF2018}, OpenVINS \cite{OpenVins2020}, and SchurVINS \cite{fan2024schurvins}.
    With advancements in nonlinear optimization theory and improvements in computational power, optimization-based VIO methods have become the dominant approach. These methods integrate observational data and system constraints by formulating optimization problems, effectively overcoming the limitations of traditional filtering methods in handling complex nonlinear systems. Notable examples include OKVIS \cite{leutenegger2015keyframe}, VINS-Mono \cite{VINS-Mono}, ORB-SLAM3 \cite{campos2021orb}, BASALT \cite{BASALT}, Kimera \cite{Kimera}, and DM-VIO \cite{DM-VIO}.
    
    However, existing methods often struggle to perform reliably under extreme conditions, such as low-texture environments or fast motion. These challenging scenarios can significantly degrade the accuracy of state estimation or even lead to system failures. Therefore, enhancing the robustness of VIO algorithms under degraded visual conditions remains a fundamental and ongoing challenge in the field.
    
    \subsection{Learning based inertial odometry}
    Learning-based inertial odometry has gained significant attention in recent years by combining traditional kinematic models with data-driven deep learning methods. One of the pioneering works, IONet \cite{chen2018ionet}, introduced a deep network-based inertial navigation system that predicts the displacement between adjacent IMU measurements, thereby mitigating error accumulation and generating relatively accurate trajectories. Building on similar tasks, Herath et al. \cite{herath2020ronin} proposed a novel neural network architecture that achieves higher localization precision, particularly in challenging motion scenarios. Further advancing this line of research, Liu et al. \cite{liu2020tlio} introduced TLIO, which leverages a neural network for 3D displacement estimation and integrates it with the EKF to produce high-fidelity trajectories. Subsequently, Sun et al. \cite{sun2021idol} proposed IDOL, a two-stage data-driven pipeline that first estimates orientation and then position, yielding improved localization results. Similarly, Zhang et al. \cite{zhang2021imu} utilized deep neural networks to compute observable IMU integration terms, enhancing the precision and robustness of inertial navigation through numerical pose integration and sensor fusion.
    
    In addition to trajectory estimation, many learning-based methods focus on IMU noise reduction and uncertainty estimation. For example, Brossard et al. \cite{brossard2020denoising} employed Convolutional Neural Networks (CNN) for gyro denoising. Steinbrener et al. \cite{steinbrener2022improved} compared different architectures for denoising IMU measurements and found that Long Short-Term Memory Networks (LSTM) outperform Transformer architectures, especially under varying data rates. AirIMU \cite{qiu2023airimu} introduced an encoder-decoder network structure to process raw IMU data, estimating IMU states and uncertainties, and outputting corrected acceleration and angular velocity values along with their associated uncertainties. Zeinali et al. \cite{zeinali2024imunet} proposed MobileResNet, a CNN-based lightweight block that utilizes deep and pointwise convolutions to reduce computational costs and noise interference, thereby improving navigation and localization. Moreover, Yang et al. \cite{yang2024enhancing} integrated the Left-Invariant Extended Kalman Filter with a statistical neural network, proposing a learning-based dead reckoning algorithm that relies solely on inertial measurements and exhibits robust performance under sudden changes in lighting conditions.
    
    However, existing methods still face challenges due to integration drift caused by accumulated noise and bias, which reduces the long-term accuracy of trajectory estimation. To address this issue, researchers have integrated IMU data with visual sensors and leverage deep learning to further improve both accuracy and robustness.
    
    \subsection{Learning based visual inertial odometry}
    Most of the current research on learning-based VIO focuses on the visual component, addressing various challenges faced by vision, such as blurred fields of view, sparse feature points, and so on. A smaller subset of studies focuses on utilizing learning-based methods to extract the potential information from IMU data for VIO. Danpeng Chen et al.\cite{chen2021rnin} proposed a fusion of LSTM and a tightly coupled EKF framework to leverage IMU measurements, improving the robustness and precision of position estimation in visually constrained scenarios. Russell Buchanan et al. \cite{Deep_IMU_2023} introduced the first method to infer the evolution of bias using neural networks and integrated it as unary factors into a state estimator based on factor graphs. Subsequently, Solodar et al. \cite{solodar2024vio} proposed VIO-DualProNet, which dynamically estimates the noise uncertainty of IMU data in real-time and fuses it into VINS-Mono \cite{VINS-Mono} to enhance the robustness and precision of the system.
    
    Unlike existing methods, we restructured the IMU bias prediction paradigm and proposed an end-to-end average bias regression network (see Section III-B for details). Based on this, we designed a regularization prior factor that can be easily integrated into any VIO method.
    
    \begin{figure*}[htbp]
        \centering
        \includegraphics[width=1.0\textwidth]{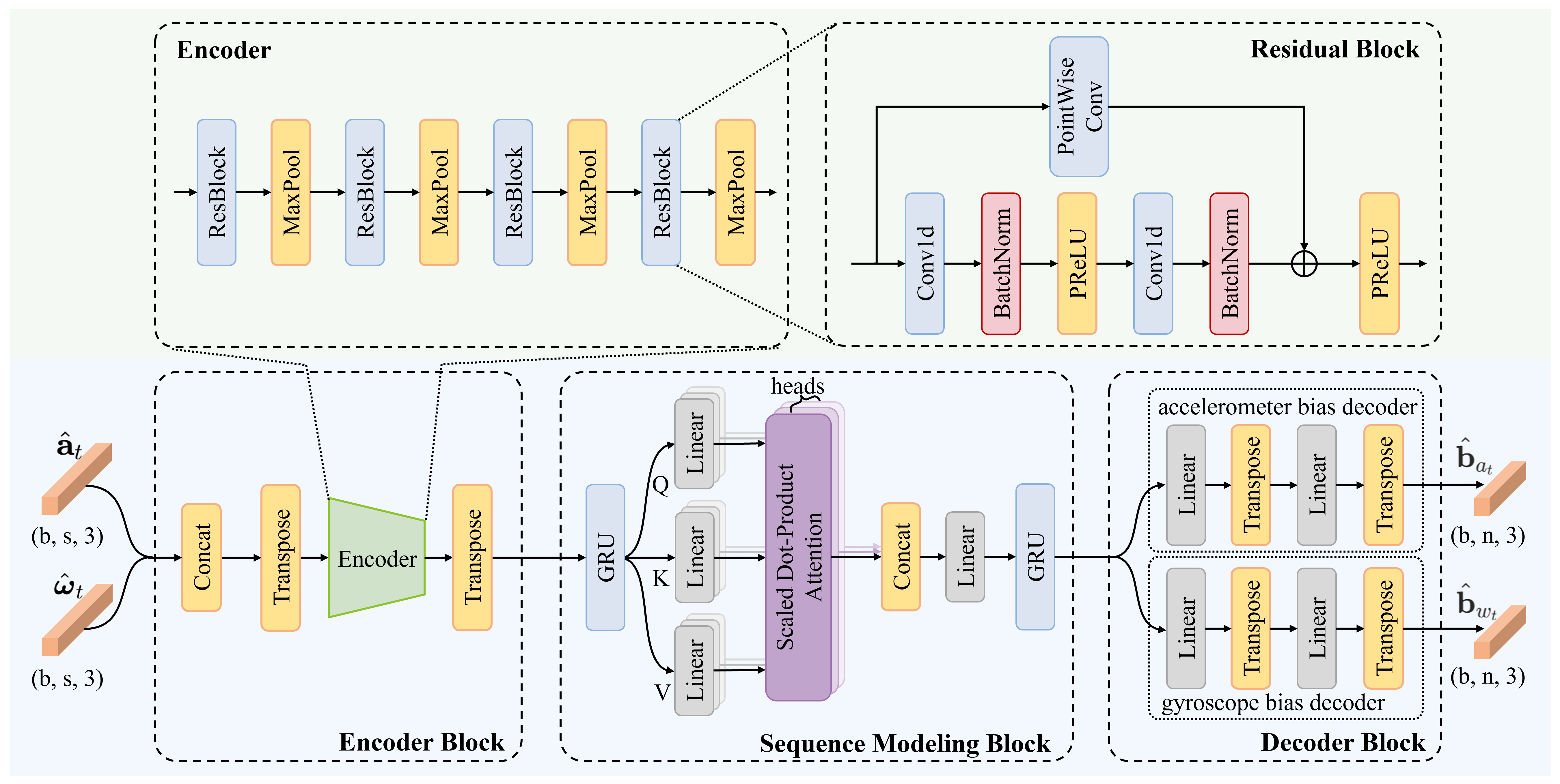} 
        \caption{The architecture of IPNet consists of encoder block, sequence modeling block and decoder block. The encoder block is mainly composed of four residual blocks and max-pooling layers, used to extract features and reduce the data dimension. The sequence modeling block is designed to further capture temporal dependencies, and the decoder block is responsible for performing regression to predict the bias of both acceleration and angular velocity.}
        \label{IPNet}
    \end{figure*}
    
    \section{Methodology}
    The overview of our proposed framework is shown in Fig. \ref{frame}. We use VINS-Fusion \cite{qin2019general} as the baseline and integrate the plug-and-play learning-based IMU bias prior factor into it. This integration addresses a key challenge in VIO frameworks: the degradation of localization precision and robustness caused by inaccurate bias estimation due to visual tracking errors. Furthermore, to facilitate the acquisition of IMU bias labels for network training, we introduce a method for iteratively solving the bias using the ground truth pose and velocity.
    
    \subsection{Preliminary}
        \subsubsection{IMU noise model}
        Take $\mathbf{a}_t$ and $\boldsymbol{\omega}_t$ as the true values of the IMU, and $\hat{\mathbf{a}}_t$ and $\hat{\boldsymbol{\omega}}_t$ as the measured values, both expressed in the body coordinate system. Considering the effects of bias, Gaussian white noise, and the gravitational acceleration in the world coordinate system, the IMU noise model can be expressed as:
            \begin{equation}
            \begin{aligned}
                \hat{\mathbf{a}}_t &=\mathbf{a}_t+\mathbf{b}_{a_t}+\mathbf{R}_w^t \mathbf{g}^w+\mathbf{n}_a, \\
                \hat{\boldsymbol{\omega}}_t &=\boldsymbol{\omega}_t+\mathbf{b}_{w_t}+\mathbf{n}_w.
            \end{aligned}
            \end{equation}
        where $\mathbf{g}^w$ represents the gravitational force, $\mathbf{R}_w^t$ stands for the rotation matrix between the world frame and the body frame, $\mathbf{b}_{a_t}$ and $\mathbf{b}_{w_t}$ denote the accelerometer bias and gyroscope bias at time $t$, respectively, expressed in the body frame, and $\mathbf{n}_a$ and $\mathbf{n}_w$ describe the accelerometer noise and angular velocity noise.
    
        \subsubsection{Preintegration}
        We follow the discrete expressions in \cite{Preintegration} and express the displacement, velocity, and rotational preintegration terms as $\boldsymbol{\alpha}_{b_{k+1}}^{b_k}$,$\boldsymbol{\beta}_{b_{k+1}}^{b_k}$,$\gamma_{b_{k+1}}^{b_k}$.
        \begin{equation}
        \begin{aligned}
            \boldsymbol{\alpha}_{b_{k+1}}^{b_k} & =\iint_{t \in\left[t_k, t_{k+1}\right]} \mathbf{R}_t^{b_k}\left(\hat{\mathbf{a}}_t-\mathbf{b}_{a_t}-\mathbf{n}_a\right) d t^2, \\
            \boldsymbol{\beta}_{b_{k+1}}^{b_k} & =\int_{t \in\left[t_k, t_{k+1}\right]} \mathbf{R}_t^{b_k}\left(\hat{\mathbf{a}}_t-\mathbf{b}_{a_t}-\mathbf{n}_a\right) d t, \\
            \boldsymbol{\gamma}_{b_{k+1}}^{b_k} & =\int_{t \in\left[t_k, t_{k+1}\right]} \frac{1}{2} \boldsymbol{\Omega}\left(\hat{\boldsymbol{\omega}}_t-\mathbf{b}_{w_t}-\mathbf{n}_w\right) \boldsymbol{\gamma}_t^{b_k} d t.
        \end{aligned}
        \end{equation}

        where:
        \begin{equation}
            \resizebox{0.85\hsize}{!}{$
                \boldsymbol{\Omega}(\boldsymbol{\omega})=\left[\begin{array}{ll}
                -\lfloor\boldsymbol{\omega}\rfloor_{\times} & \boldsymbol{\omega} \\
                -\boldsymbol{\omega}^T & 0
                \end{array}\right], \lfloor\boldsymbol{\omega}\rfloor_{\times}=\left[\begin{array}{ccc}
                0 & -\omega_z & \omega_y \\
                \omega_z & 0 & -\omega_x \\
                -\omega_y & \omega_x & 0
                \end{array}\right].$}
        \end{equation}
    
    \subsection{IPNet model}
    In this work, we propose IPNet to directly predict the bias from raw IMU data according to the motion prior learned from existing trajectories \cite{Deep_IMU_2023}. The network consists of three blocks: encoder block, sequence modeling block, and decoder block, as shown in Fig.\ref{IPNet}. The raw IMU data represent the acceleration and angular velocity measurements in three dimensions, and we take a sample window of length $s$ as the input to the model, so that the measurements can be denoted as $\hat{\mathbf{a}}_t \in \mathbb{R}^{s \times 3}$ and $\hat{\boldsymbol{\omega}}_t \in \mathbb{R}^{s \times 3}$. We predict $n$ values for each window to avoid prediction fluctuations. Therefore, the model outputs the predicted bias $\hat{\mathbf{b}}_a \in \mathbb{R}^{n \times 3}$ and $\hat{\mathbf{b}}_w \in \mathbb{R}^{n \times 3}$. In this paper, we set $s$ to 1000 and $n$ to 50.
    
    \subsubsection{Encoder block}
    The encoder block mainly consists of several transformation layers to reshape the tensor and an encoder to capture local features and reduce data dimension. Firstly, concatenate $\hat{\mathbf{a}}_t$ and $\hat{\boldsymbol{\omega}}_t$ along the last dimension, and then use the transpose operation to swap the positions of the last two dimensions, so as to facilitate the subsequent convolution operations. The encoder contains four sets of residual blocks and max-pooling layers. The main path of residual block primarily utilizes 1D convolution, batch normalization, and PReLU activation, and the residual connection employs a pointwise convolution layer to combine the features of different channels and make the input and output have the same dimension, which are then added to the main path. Finally, a transpose operation is applied to restore the original dimensional order.  
    
    \subsubsection{Sequence modeling block}
    The sequence modeling block is used to learn the temporal dependencies from the features output by the encoder, which consists of two Gated Recurrent Unit (GRU) layers with a multi-head self-attention\cite{vaswani2017attention} inserted between them. GRU is a kind of gated recurrent neural network. It controls the flow of information by introducing the update gate and the reset gate, thus realizing the modeling of long-term dependencies. The multi-head self-attention mechanism is used to further capture the relationships among different positions in the feature sequence, so that the model can focus on different context information in different semantic subspaces through multiple linear transformations.
    
    \subsubsection{Decoder block}
    To accommodate the characteristics of acceleration bias and angular velocity bias, two parallel decoders predict accelerometer and gyroscope biases. Additionally, to avoid instability caused by fluctuations in model predictions, each decoder outputs $n$ predicted values, and their mean is used as the final prediction for each data window.  
    
    \subsubsection{Loss function}
    We use the Mean Absolute Error (MAE), to measure the difference between the predicted values and the ground truth. This loss pays more attention to the performance of the model on the majority of samples and is more robust to outliers. The final loss $L$ is obtained by the sum of the L1 losses of the acceleration bias and the angular velocity bias. The formula is as follows: 
    \begin{equation}
    L=L_{1_{b_a}}+L_{1_{b_w}}=\frac{1}{N}\sum_{t=1}^N|\mathbf{b}_{a_t}-\hat{\mathbf{b}}_{a_t}|+\frac{1}{N}\sum_{t=1}^N|\mathbf{b}_{w_t}-\hat{\mathbf{b}}_{w_t}|.
    \end{equation}
    
    \begin{figure*}[htbp]
        \centering
        \includegraphics[width=1.0\textwidth]{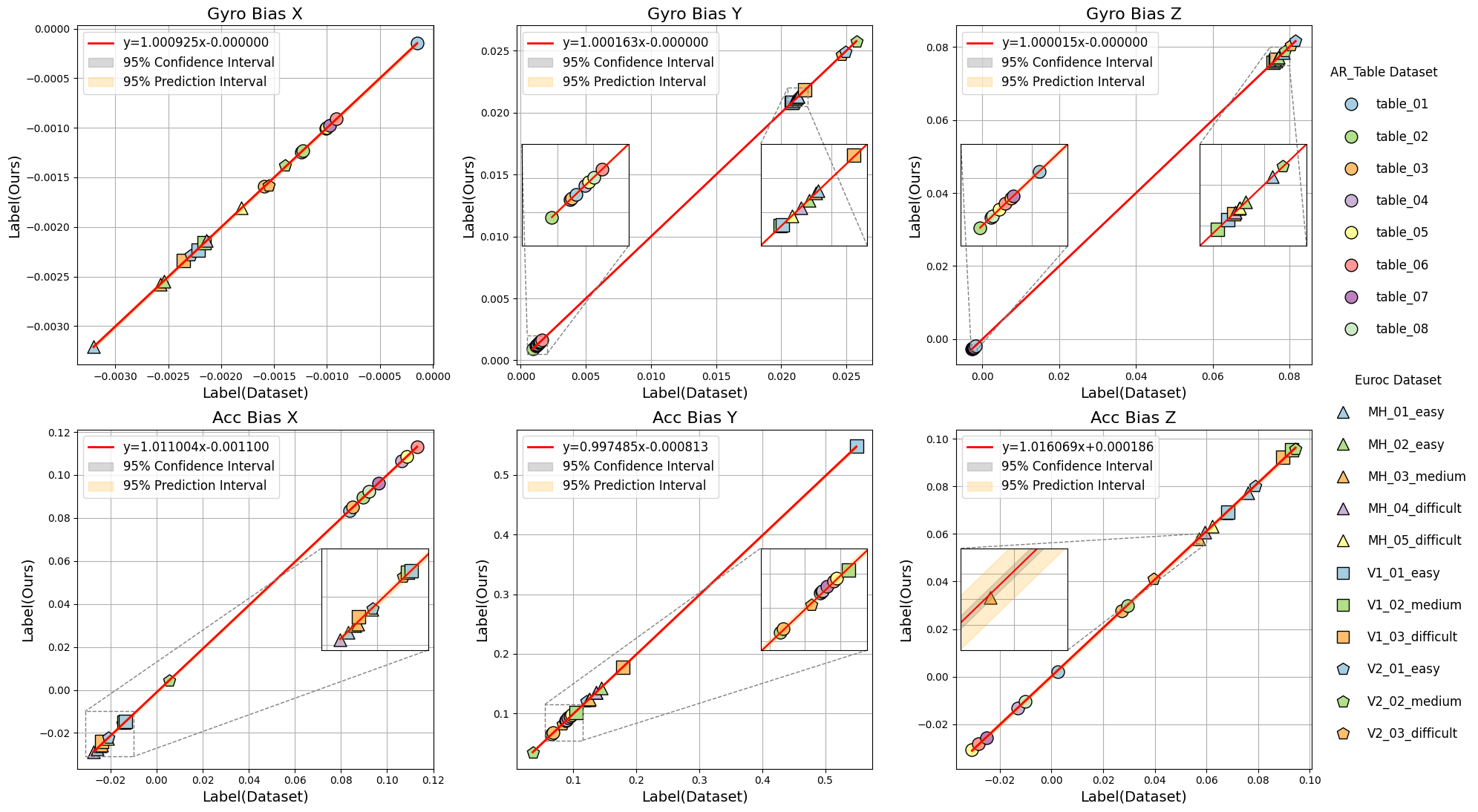}
        \caption{IMU Bias Calculation Accuracy Experiment. The horizontal axis represents the mean of the ground truth bias, and the vertical axis represents the bias label calculated by our method. Ideally, they should be equal and distributed along a straight line with a slope of 1 and passing through the origin. The figure displays the bias distribution and fitting of different datasets. The gray area represents the confidence interval, used to characterize the 95\% confidence interval of the fitting curve, while the yellow area represents the prediction interval, used to characterize the estimated range of future observation values under the 95\% confidence level.}
        \label{IMU_Bias_Comp}
    \end{figure*}
    
    \subsection{IMU bias prior factor using IPNet}
    In this work, we integrate IPNet into any VIO system by forming an IMU bias factor using the IPNet. The main purpose of this factor is to provide a bounded value for the bias in the optimization procedure within the backend of VIO. It evaluates the deviation between the actual bias value and the known initial value in the form of residual, and incorporates this constraint into the optimization problem, helping to reduce state drift. The definition of its residual is:
        \begin{equation}
        \mathbf{r}={\mathbf{W}} \cdot\left[\begin{array}{l}
        \mathbf{r}_{b_a} \\
        \mathbf{r}_{b_w}
        \end{array}\right]={\mathbf{W}} \cdot\left[\begin{array}{l}
        \mathbf{b}_{a_t, \text { p }}-\hat{\mathbf{b}}_{a_t} \\
        \mathbf{b}_{w_t, \text { p }}-\hat{\mathbf{b}}_{w_t}
        \end{array}\right].
        \end{equation}
    where $\mathbf{b}_{a_t, \text{p}}$ and $\mathbf{b}_{w_t, \text{p}}$ represent the final optimized results of the accelerometer bias and angular velocity bias at time $t$, respectively. The term $\mathbf{W}$ represents the weight matrix, which controls the influence of the residuals on the optimization objective. Its Jacobian matrix is:
        \begin{equation}
        \mathbf{J_r}={\mathbf{W}} \cdot\left[\begin{array}{ll}
        \mathbf{I}_{3 \times 3} & \mathbf{0}_{3 \times 3} \\
        \mathbf{0}_{3 \times 3} & \mathbf{I}_{3 \times 3}
        \end{array}\right].
        \end{equation}  
    
    \subsection{IMU bias estimation for training label}
    A fundamental challenge in learning IMU bias priors is the absence of ground-truth bias values in most visual-inertial datasets. To address this, we propose an optimization-based method to compute per-sequence average biases, which serve as the supervision signal for IPNet. This design choice is motivated by a key observation: while IMU biases theoretically undergo random walk, their variation over a typical VIO sequence is often dominated by a relatively stable offset. Our solution proceeds as follows: first, we leverage the pose ground truth and raw IMU measurements to iteratively solve for the IMU bias that best aligns the pre-integrated trajectory with the ground truth. Then, we average these biases over the entire sequence to obtain a single, representative label. By regressing towards this sequence average, IPNet learns to predict a stable and representative bias prior that captures the dominant error component induced by the specific platform's motion pattern, rather than attempting to track the noisy instantaneous bias. This approach effectively smoothes out the high-frequency noise inherent in the optimization-based inverse calculation and provides a robust learning target for the network.
    
    The steps of the algorithm are as follows: First, the raw IMU data is pre-integrated to obtain the measured values of the pre-integrated terms. These values are then subtracted from the ground truth of the pre-integrated terms to calculate the error. Finally, $\mathbf{b}_{a_t}$ and $\mathbf{b}_{w_t}$ are inversely solved through iterative calculations based on the first-order term of the pre-integrated with respect to bias changes. The formula is as follows:
    \begin{equation}
    \begin{aligned}
    \alpha_{b_{k+1}}^{b_k} - \hat{\alpha}_{b_{k+1}}^{b_k} &\approx J_{b_a}^\alpha \delta b_a+J_{b_\omega}^\alpha \delta b_\omega, \\
    \beta_{b_{k+1}}^{b_k} - \hat{\beta}_{b_{k+1}}^{b_k} &\approx J_{b_a}^\beta \delta b_a+J_{b_\omega}^\beta \delta b_\omega, \\
    \hat{\gamma}_{b_{k+1}}^{b_k}{}^{-1} \cdot \gamma_{b_{k+1}}^{b_k} &\approx \left[\frac{1}{2} J_{b_\omega}^\gamma \delta b_\omega\right].
    \end{aligned}
    \end{equation}
    where the specific form of $J$ refers to \cite{VINS-Mono}. Assuming the initial values of $b_{a_{init}}$ and $b_{w_{init}}$ are 0, a linear expansion at this point can be used to calculate the bias at each moment as $\mathbf{b}{a_t} = \delta b_a + b_{a_{init}}$ and $\mathbf{b}{w_t} = \delta b_w + b_{w_{init}}$. To directly obtain the mean values of $\mathbf{b}{a_t}$ and $\mathbf{b}{w_t}$, a similarity constraint can be further applied to the entire sequence, iteratively solving for the final IMU bias. These biases can then be used as the training labels for the network.
    
    
    \section{Experiments}
    Two public datasets with IMU bias labels were used to verify the accuracy of our iterative method and its resulting IMU bias labels. The resulting label data is released publicly. Then, we evaluated the proposed framework on public datasets and In-house dataset. Localization precision and robustness were validated against the VINS-Fusion\cite{qin2019general} Mono-IMU baseline across two public datasets. Finally, performance assessment was also conducted on In-house dataset, followed by generalization testing through cross-domain transfer of indoor-trained models to outdoor environments.
    
    \subsection{Accuracy evaluation of IMU bias calculations in training set and training details}
    We validate the accuracy of our iterative method on the EuRoC and AR\_Table datasets by comparing the calculated results with the ground truth average. The comparison results are shown in Fig.\ref{IMU_Bias_Comp}. The slope and intercept of the curves in the figure clearly reflect the calculation accuracy, which can be directly used as labels for network training.
    
    \subsection{Experimental setup}
    
    \begin{figure}[htbp]
        \centering
        \includegraphics[width=0.48\textwidth]{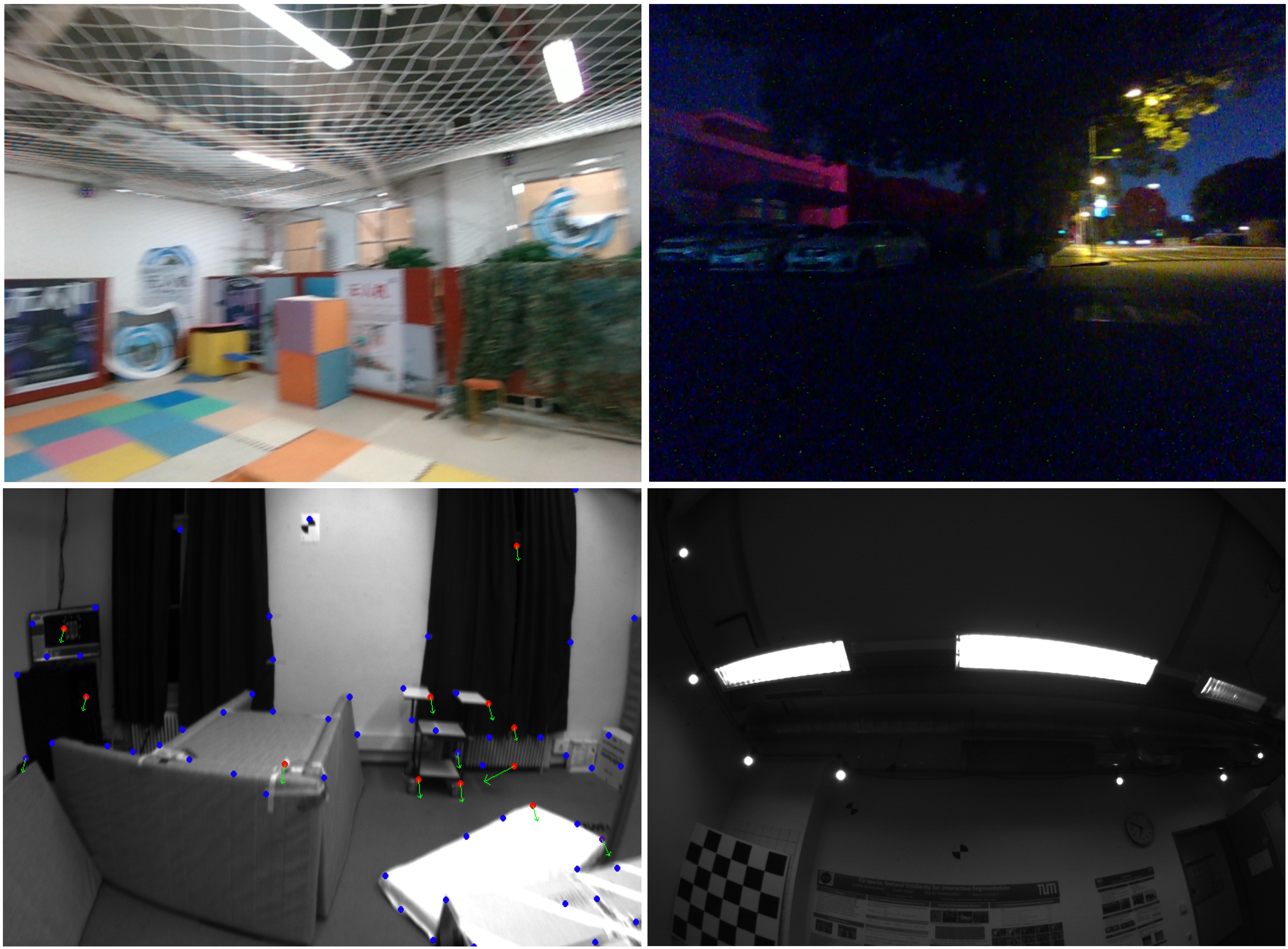}
        \caption{The image shows schematic diagrams of different scenes in challenging environments, which correspond to the following test sequences: (a): the Seq04 sequence of the In-house dataset, (b): the Seq17 sequence of the In-house dataset, (c): the V2\_02 sequence of the EuRoC dataset, and (d): the room\_5 sequence of the TumVi dataset. Due to poor visual features in these scenes, the trajectories diverge, affecting the robustness of the system.}
        \label{scene}
    \end{figure}
    
    \subsubsection{Datasets}We select the EuRoC \cite{burri2016euroc}, TumVi \cite{schubert2018tum}, AR\_Table \cite{chen2023monocular} and In-house dataset for the evaluation.
    
    \textbf{EuRoC:} This dataset is widely used and consists of 20Hz drone images and 200Hz inertial data, with accurate the ground truth pose obtained through a motion capture system. Fortunately, each sub-dataset provides the ground truth of IMU bias, offering a valuable reference for verifying the effectiveness of our bias-solving method.
    
    \textbf{TumVi:} This dataset was collected via handheld platforms across diverse scenarios, exhibiting substantially different motion patterns from the EuRoC's smooth aerial maneuvers, thus serving as an established benchmark for evaluating VIO system robustness.
    
    \textbf{AR\_Table:} This dataset is primarily designed for 3D reconstruction, providing motion-capture ground truth poses and biases similar to EuRoC. Due to its visually rich scenes diminishing the reliance on IMU constraints, it is exclusively used for evaluating IMU bias estimation accuracy.
    
\begin{figure}[htbp]
        \centering
        \includegraphics[width=0.48\textwidth]{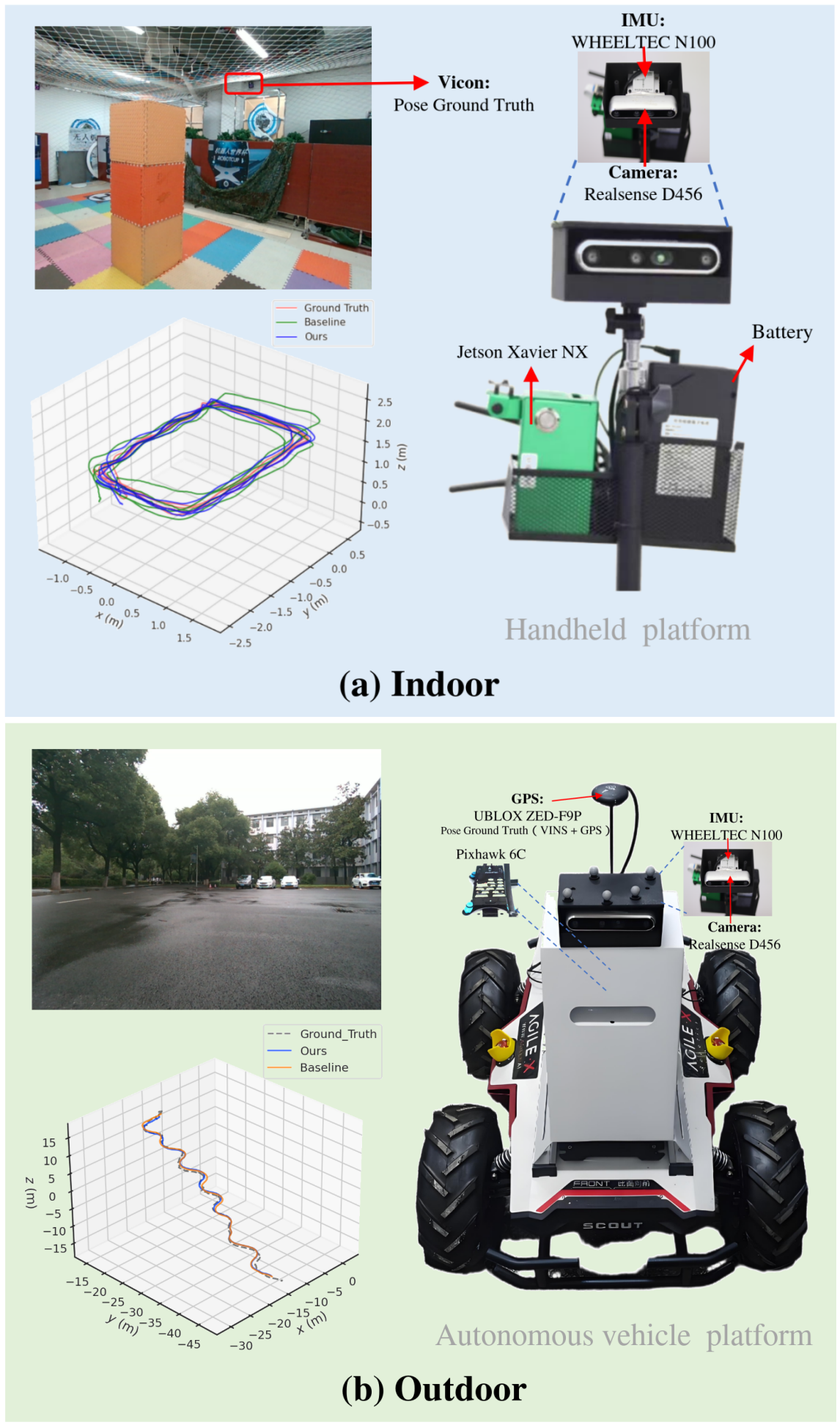}
        \caption{The image respectively illustrates the environmental setup, equipment platform, and partial trajectory comparison results obtained during the data collection process. The top trajectory corresponds to the Seq02 sequence of the in-house dataset, while the bottom trajectory corresponds to the Seq16 sequence of the same dataset.}
        \label{traj}
    \end{figure}

    \textbf{In-house Dataset:} We built a modular visual-inertial sensor unit using the low-cost WHEELTEC N100 IMU (200Hz) and Realsense D456 camera, as shown in Fig. \ref{traj}. Initially, the unit was integrated into a handheld device, and 12 motion sequences were collected indoors (total: 20 minutes). Since only one side of the Vicon system was functional, we focused on planar motion to ensure reliable pose ground truth. The unit was then transferred to an autonomous vehicle platform, where 5 additional sequences were collected outdoors (total: 8 minutes), with pose ground truth obtained using the GPS mode in VINS-Fusion \cite{qin2019general}. The GPS module we use is the UBLOX ZED-F9P, with an RTK positioning accuracy of 0.01m + 1ppm CEP.
    
    Fig. \ref{scene} shows several challenging scenes from the dataset. Due to issues such as motion blur, sparse textures, and low lighting, the visual features are poor for a common VIO system. As a result, the baseline system fails to localize in these scenes. 
    
    \subsubsection{Evaluation metrics}
    Localization precision is quantified using absolute trajectory error (ATE-RMSE/M) and relative pose error (RPE-RMSE/RAD) per sub-dataset. System robustness is characterized visually through trajectory divergence/convergence phenomena. IMU bias accuracy is validated by comparing iterative solutions against per-dimension bias means.
    
    \subsubsection{Implementation details}
    Experiments were conducted on a Lenovo Y9000P platform equipped with an RTX 4090 GPU, running Ubuntu 20.04 with OpenCV 3.4.5.
    
    We implemented the proposed IPNet using the PyTorch framework, with MAE as the loss function and RMSprop as the optimizer. The initial learning rate is set to 1e-6 and decays by a factor of 10 every 10 epochs. Statistical analysis of the sample distribution revealed significant imbalance across sub-datasets. To address this, we adopted a training strategy different from previous studies \cite{brossard2020denoising, Deep_IMU_2023, qiu2023airimu}.The dataset splits are as follows: for the EuRoC dataset, MH\_01, MH\_03, MH\_04, V1\_03, and V2\_03 are used for training, MH\_02 and V1\_01 for validation, and the remaining sequences for testing; for TumVi, room1, room2, and room4 are used for training, room3 for validation, and the remaining two sequences for testing; for In-house Dataset, six indoor sequences are used for training, two indoor sequences for validation, and the remaining four indoor sequences together with all outdoor sequences form the test set. Finally, to ensure a fair comparison with previous recursive methods, we conducted additional experiments by retraining those models under the same data splitting strategy for evaluation.
    
    \begin{table}[htbp]
        \centering
        \caption{ATE [M] \& RPE [RAD]: Performance on public datasets}
        \setlength\tabcolsep{3pt}
        \label{tab-001}
        \begin{tabular}{l c c c c c}
            \toprule
                \multicolumn{2}{c}{\multirow{2}{*}{Sequence}} &\multicolumn{2}{c}{Baseline} & \multicolumn{2}{c}{Ours} \\
                \cmidrule(lr){3-4} 
                \cmidrule(lr){5-6}
                    &        & ATE     & RPE    &   ATE &     RPE \\
                \midrule
                \multicolumn{1}{c}{\multirow{4}{*}{EuRoC}} 
                    & MH\_05 & 0.340   & 0.02115 & 0.249(27\%) & 0.00655(69\%)  \\
                    & V1\_02 & 0.270   & 0.03022 & 0.106(61\%) & 0.00754(75\%)  \\
                    & V2\_01 & 0.133   & 0.04317 & 0.048(64\%) & 0.00672(84\%)  \\
                    & V2\_02 & -       & -      & 0.098$^*$   & 0.00565$^*$    \\
                \midrule
                \multicolumn{1}{c}{\multirow{2}{*}{TumVi}} 
                    & room5 & -        & -      & 0.436$^*$   & 0.10512$^*$    \\
                    & room6 & 0.198    & 0.03645 & 0.102(48\%) & 0.04145(-14\%) \\
                    \midrule
                    \multicolumn{2}{c}{Average} & 0.235 & 0.0327 & 0.126(\textbf{46\%}) & 0.0157(\textbf{48\%}) \\
            \toprule
            \multicolumn{6}{l}{\footnotesize{$^*$ Do not participate in the calculation of the average.}}\\
            \multicolumn{6}{l}{\footnotesize{- indicates trajectory divergence of the baseline.}}\\
        \end{tabular}
    \end{table}	
    
    For the calculation of IMU bias labels, we first optimize the gyroscope bias using the Adam optimizer with an initial learning rate of 0.001 for 15,000 iterations, reducing the learning rate by a factor of 10 every 5,000 iterations. The resulting gyroscope bias is then fixed, and the accelerometer bias is optimized with a learning rate of 0.01, while keeping all other hyperparameters unchanged.
    \subsection{Performance}
    
    \begin{figure}[htbp]
        \centering
        \includegraphics[width=0.48\textwidth]{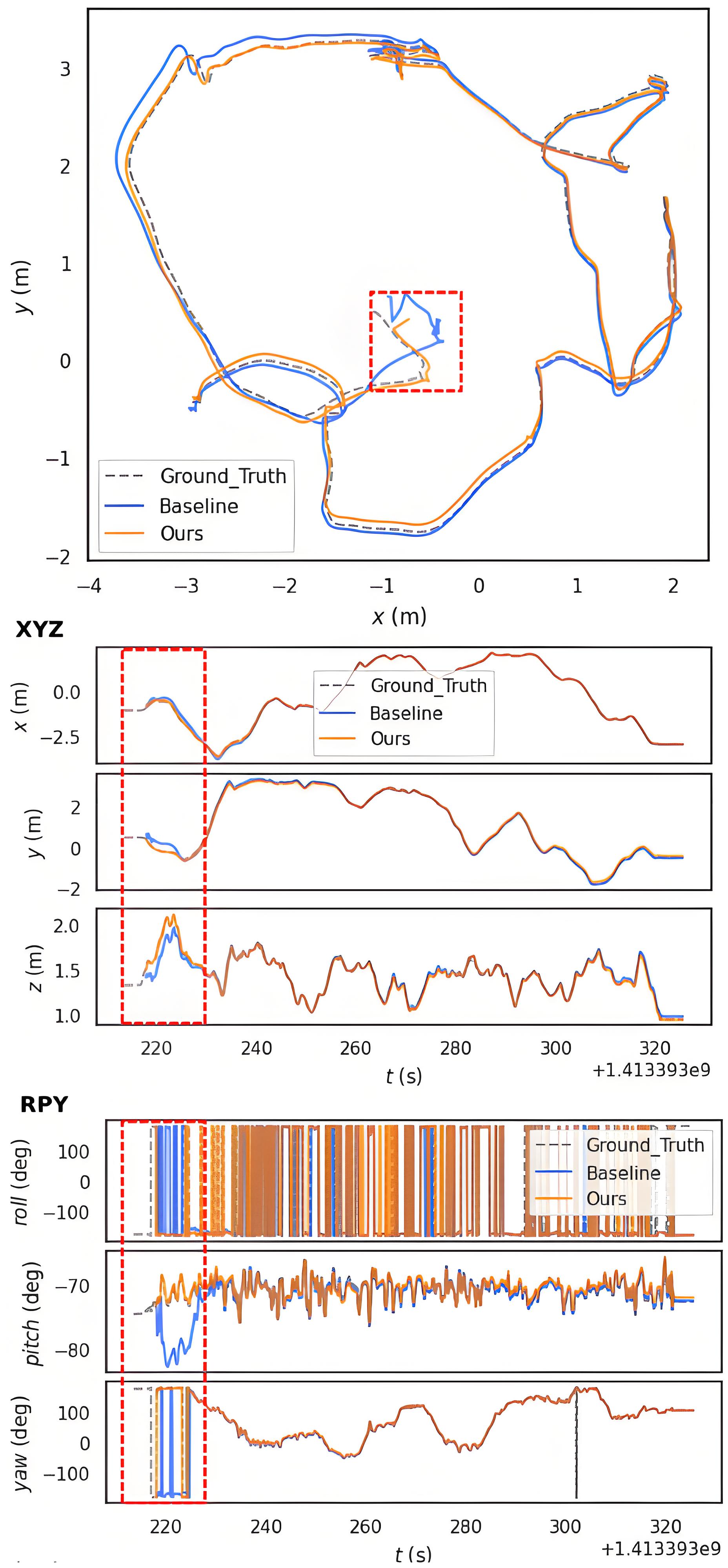} 
        \caption{The image shows the trajectory and pose comparison between our method and the baseline for the V2\_01 sequence of the EuRoC dataset, with the red box highlighting the comparison during the initial phase.}
        \label{exp5}
    \end{figure}
    
    \begin{figure*}[htbp]
        \centering
        \includegraphics[width=1.0\textwidth]{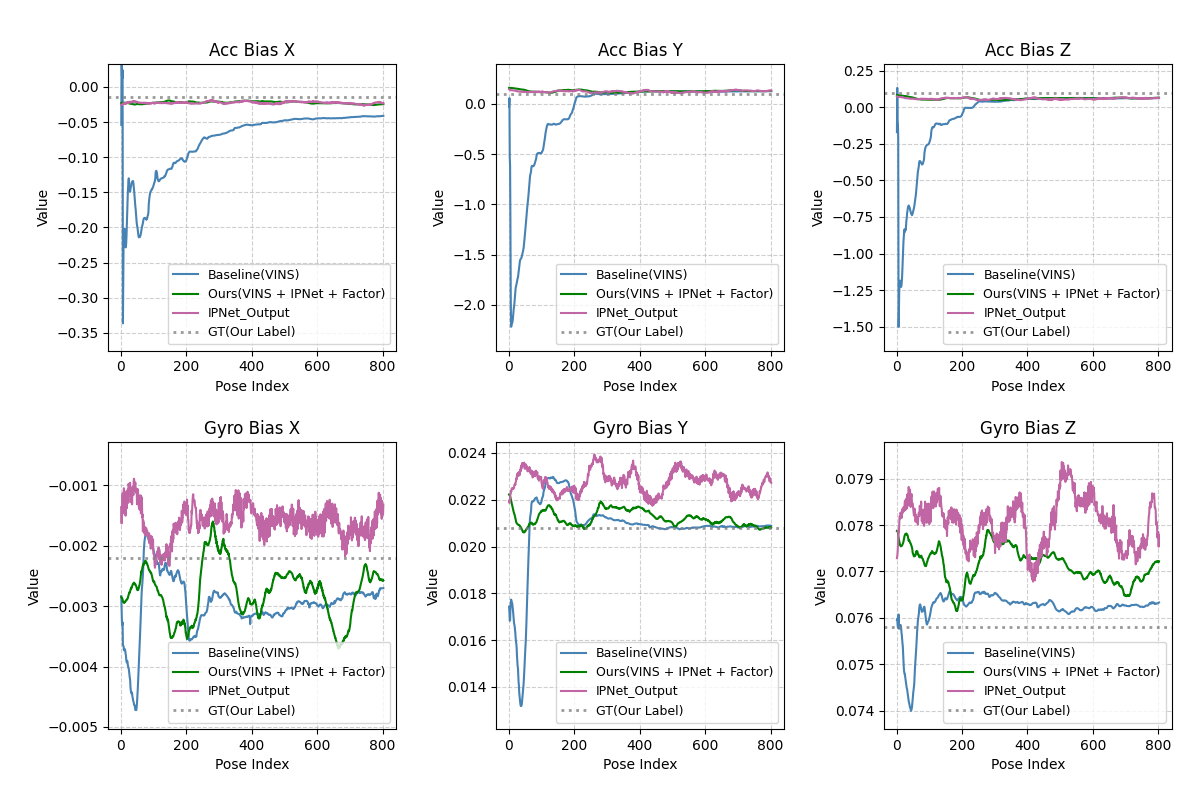} 
        \caption{The comparison of bias values curves corresponding to each pose in the V1\_02 sequence of the EuRoC dataset. In the baseline, the bias values corresponding to the first 200 poses deviate significantly from the ground truth, whereas with our method, the bias values are quickly adjusted to fluctuate close to the ground truth. Pink: IPNet output; Blue: the bias predicted by baseline; Green: the bias predicted by our framework; Red: our label.}
        \label{Bias_perf}
    \end{figure*}
    
    As shown in Table \ref{tab-001}, the proposed framework significantly enhances localization precision on public datasets, with ATE-RMSE and RPE-RMSE improved by an average of 46\% and 48\%, respectively. The baseline method experiences trajectory divergence in sequences V2\_02 and room5, mainly due to large visual tracking errors and a low image processing rate of only 10Hz, which leads to insufficient inertial constraints for error compensation. In the case of room5, frequent shaking causes drastic changes in viewpoint, and the minimum distance threshold in visual matching limits the number of usable feature points at certain moments, eventually resulting in divergence. In contrast, our method introduces an IMU bias prior, providing more accurate bias constraints to the system, thereby significantly enhancing localization accuracy and robustness even under severe visual degradation. It is worth noting that in the room6 sequence, the RPE-RMSE slightly decreased, possibly due to the limited training data and the simplicity of the motion pattern, which prevented the network from learning more complex motion patterns.
    
    During the VIO initialization phase, the system is often affected by poor visual features and severe vibrations, which cause the estimated IMU bias to deviate significantly from the actual physical value, thus severely affecting the accuracy of the initial pose. This deviation can lead to a decline in the overall performance. As shown in Fig. \ref{exp5}, during the initialization phase, our approach applies constraints during the IMU bias estimation process, effectively limiting its deviation and ensuring the accuracy of both IMU bias and pose estimation.
    
    Fig. \ref{Bias_perf} shows the pose bias variations during VIO tracking for our method versus the baseline on sequence V1\_02. By incorporating prior, biases are effectively constrained within reasonable bounds, enhancing estimation accuracy and system stability. While the gyroscope's Z-axis estimates are marginally inferior to the baseline due to shifted optimization priorities following prior introduction, their numerical discrepancy remains minimal.
    
    \begin{table}[h]
        \centering
        \caption{ATE [M] \& RPE [RAD]: Performance on EuRoC dataset}
        \setlength\tabcolsep{3pt}
        \label{tab-002}
        \begin{tabular}{l c c c c c c}
            \toprule
            \multirow{2}{*}{Sequence} & \multicolumn{2}{c}{Zhang et al.\cite{zhang2021imu}}    & \multicolumn{2}{c}{Russell B. et al.\cite{Deep_IMU_2023}} & \multicolumn{2}{c}{Ours} \\
                \cmidrule(lr){2-3} 
                \cmidrule(lr){4-5}
                \cmidrule(lr){6-7}
                    &       ATE     & RPE    &   ATE &     RPE &   ATE &     RPE\\
                \midrule
                MH\_02 & 0.150          & \textbf{0.0025} & 0.130 & - & \textbf{0.087} & 0.0026 \\
                MH\_04 & \textbf{0.138} & \textbf{0.0037} & 0.250 & - & 0.188 & 0.0072 \\
                V1\_01 & 0.066          & 0.0162 & 0.080 & - & \textbf{0.053} & \textbf{0.0056} \\
                V1\_03 & \textbf{0.147} & \textbf{0.0067} & 0.170 & - & \textbf{0.147} & 0.0110 \\
                V2\_02 & 0.104          & 0.0068 & 0.100 & - & \textbf{0.081} & \textbf{0.0044} \\
            \toprule
            \multicolumn{6}{l}{\footnotesize{- RPE not reported in the original paper.}}\\
        \end{tabular}
        \vspace{-10pt}
    \end{table}
    
    To facilitate direct comparison with recursive methods, we retrained and evaluated our model using identical training protocols (Table \ref{tab-002}). While our approach demonstrates measurable advantages, fundamental architectural disparities preclude strictly equivalent comparison. Nevertheless, the experimental results show that the non-recursive method for learning a stable bias prior is highly competitive with the recursive time-varying bias estimation framework.
    
    \begin{table}[htbp]
        \centering	
        \setlength\tabcolsep{3pt}
        \caption{ATE [M] \& RPE [RAD]: Performance on In-house Dataset}\label{tab-003}
        \begin{tabular}{l c c c c c}
            \toprule
                \multicolumn{2}{c}{\multirow{2}{*}{Sequence}} &\multicolumn{2}{c}{Baseline} & \multicolumn{2}{c}{Ours} \\
                \cmidrule(lr){3-4} 
                \cmidrule(lr){5-6}
                &        & ATE     & RPE    & ATE   & RPE \\
                \midrule
                \multicolumn{1}{c}{\multirow{4}{*}{Indoor}} 
                & Seq01 & 0.2158 & 0.056 & 0.1641(24\%) & 0.056(0\%)  \\
                & Seq02 & 0.2844 & 0.052 & 0.0979(66\%) & 0.049(6\%)  \\
                & Seq03 & 0.1390 & 0.095 & 0.0806(42\%) & 0.094(1\%)  \\
                & Seq04 & -      & -     & 0.5578$^*$  & 0.096$^*$    \\
                \midrule
                \multicolumn{1}{c}{\multirow{5}{*}{Outdoor}} 
                & Seq13 & 0.9575 & 0.115 & 0.6486(32\%) & 0.118(-3\%) \\
                & Seq14 & 1.6255 & 0.128 & 1.5218(6\%) & 0.124(3\%) \\
                & Seq15\textsuperscript{†} & 0.4150 & 0.165 & 0.3990(4\%) & 0.166(0\%) \\
                & Seq16\textsuperscript{†} & 0.8047 & 0.179 & 0.6202(23\%) & 0.175(2\%) \\
                & Seq17\textsuperscript{†} & -      & -     & \checkmark & \checkmark \\
                \midrule
                \multicolumn{2}{c}{Average} & 0.635 & 0.113 & 0.505(\textbf{20\%}) & 0.112(1\%)   \\
            \toprule
            \multicolumn{6}{l}{\footnotesize{$^*$ Do not participate in the calculation of the average.}}\\
            \multicolumn{6}{l}{\footnotesize{\textsuperscript{†} indicates that the sequence was collected in the evening.}} \\
            \multicolumn{6}{l}{\footnotesize{\checkmark indicates that the trajectory converges(lacking ground truth).}} \\
            \multicolumn{6}{l}{\footnotesize{- indicates trajectory divergence of the baseline.}}\\
        \end{tabular}
        \vspace{-10pt}
    \end{table}
    
    Tab. \ref{tab-003} shows the comparison results between our method and the baseline method on the self-collected dataset. In terms of localization accuracy, the proposed method improves the ATE-RMSE by an average of 20\% compared to the baseline, with a more significant improvement in indoor scenes than outdoor ones, which aligns with expectations. However, it did not show a significant advantage in the RPE-RMSE metric. This can be attributed to two main factors: first, the limited coverage of the Vicon system in indoor data collection, which focused mainly on planar motion, leading to insufficient learning of roll and pitch modeling; second, the presence of dynamic objects, such as pedestrians and vehicles, in outdoor scenes, which affected pose estimation accuracy. For Seq17, we deliberately selected an evening scene with significant visual errors. Due to the divergence of the baseline trajectory, accurate pose ground truth could not be obtained, and thus, quantitative results cannot be provided. Overall, while improvements in RPE-RMSE are limited, the method demonstrates notable advantages in system robustness, which is consistent with our previous findings on public datasets.
    
    \section{Discussion}
    
    \subsection{Analysing of Inference Speed}
    
\begin{figure}[htbp]
        \centering
        \includegraphics[width=0.48\textwidth]{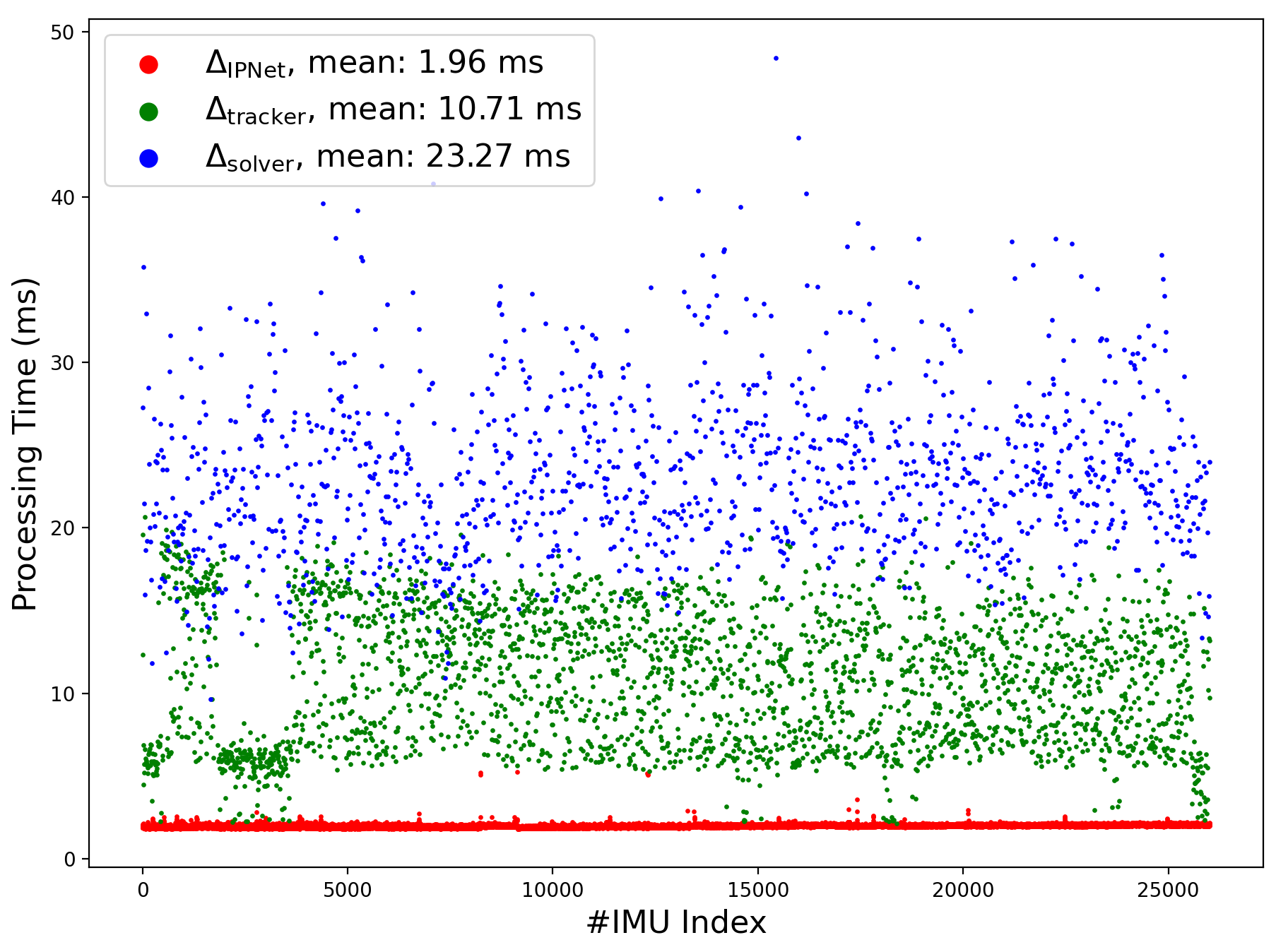}
        \caption{The time performance analysis in the MH\_03 sequence of the EuRoC dataset.}
        \label{scatter}
    \end{figure}

    As shown in Fig. \ref{scatter}, We used a subset of the EuRoC dataset to calculate the average inference time of the network, with a total of over 25,000 inference iterations. The results show that the IPNet achieves an inference frequency of 500Hz, which is 2.5 times the IMU output frequency. Combined with the time for feature point tracking and backend optimization solving, the overall system can achieve a running frequency of 30Hz, meeting the real-time operation requirements. The entire real-time performance test was conducted on a system equipped with an RTX 4090 GPU running Ubuntu 20.04.
    
    \subsection{Analysing of Generalization of IPNet}
    
    We performed data collection using a handheld device in an indoor environment, then completed model training and evaluation. Subsequently, we applied the trained model to the outdoor autonomous vehicle equipment. As shown in the experimental results in Tab. \ref{tab-003}, although there was an improvement in positioning accuracy and robustness, and the model demonstrated good generalization ability, the improvement was slightly reduced compared to the indoor environment. The evaluation demonstrated the generalization of the proposed IPNet on different ground platforms with similar motion patterns.
    
    \subsection{Analysing the requirement of Vicon}
    In this work, we collect GT pose using Vicon for calculating the bias labels. Then, we come up with a question: can we get the valid labels without Vicon, so scale up the application of the proposed method? 
    
\begin{figure}[htbp]
        \centering
        \includegraphics[width=0.48\textwidth]{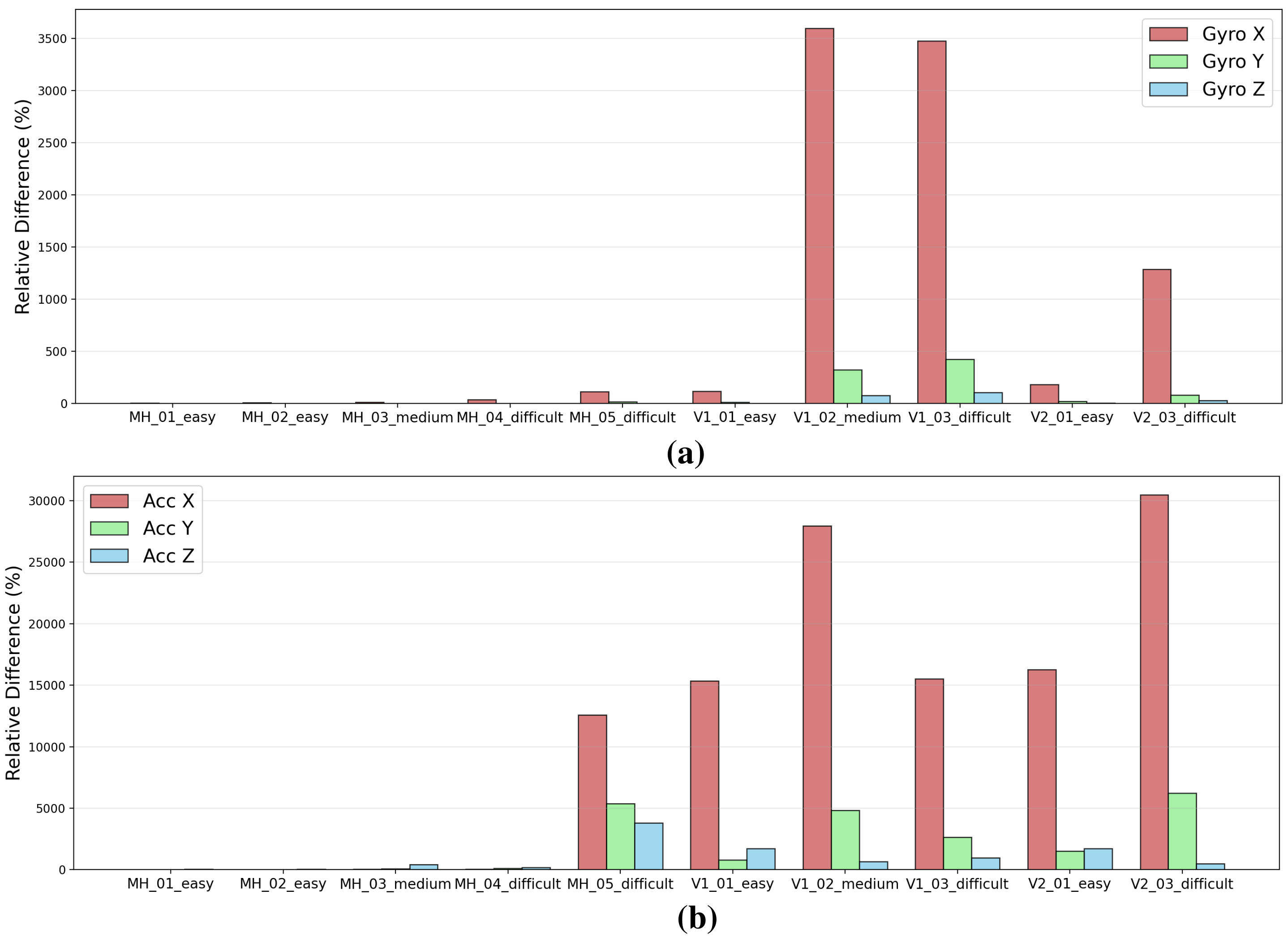}
        \caption{The difference between the results obtained by directly solving for the bias using the poses generated by VIO and those obtained using Vicon. (a): The differences in the three axes of the gyroscope, (b): The differences in the three axes of the accelerometer.}
        \label{vicon}
    \end{figure}

    To answer this question, we conducted the following experiments. Fig. \ref{vicon} shows the difference between the IMU biases directly obtained by VIO and those obtained by Vicon in the EuRoC dataset. In the MH\_01 and MH\_02 sequences, the differences are relatively small, but in other sequences, the differences become very significant due to the substantial increase in visual errors.
    
    From the experiments, we can observe that good visual features can help the VIO backend estimate the IMU bias accurately. Since our network uses the average value of the IMU bias as the supervision signal, and the average naturally smooths out noise. In the cases with good visual features, our method can obtain the IMU bias iteratively using the pose generated by VIO, without relying on the ground truth pose, and use it as the supervision signal. However, in scenarios with significant visual errors, it is necessary to rely on Vicon to obtain ground truth for iterative solving. This insight paves the way for our future work: building a scalable, self-supervised training pipeline by automatically curating high-confidence segments from large-scale VIO trajectories, effectively eliminating the dependency on external motion capture systems.
    
    \section{Conclusions}
    In this paper, we propose IPNet, a method that directly infers IMU bias prior from raw IMU data using a sliding window approach. The inferred prior is integrated into the VIO framework as plug-and-play factors. This effectively reduces IMU bias estimation errors caused by unreliable visual features, significantly improving positioning accuracy and enhancing the system's robustness. We implemented and evaluated the module on multiple datasets. Experimental results show that while the bias prior estimates output by the network are not the final precise values, they can significantly improve the overall system performance through collaborative optimization with the VIO backend. Future plans include introducing noise label optimization for training, and based on the strong robustness of the current device-specific strategy, further exploring selecting the good sequence automatically to make the proposed method scalable without Vicon.
    
\bibliographystyle{IEEEtran}
\bibliography{References}

\end{document}